\documentclass[10pt,twocolumn,letterpaper]{article}

\usepackage{cvpr}              % To produce the CAMERA-READY version
\usepackage[dvipsnames]{xcolor}

\def\paperTitle{
Hunyuan-GameCraft: High-dynamic Interactive Game Video Generation \\ with Hybrid History Condition 
}

\definecolor{cvprblue}{rgb}{0.21,0.49,0.74}
\usepackage[pagebackref,breaklinks,colorlinks,citecolor=cvprblue]{hyperref}

\newcommand{\coloneqqb}{\mathrel{\mathop:}\mathrel{\mkern-1.2mu}=} % 公式1的:=符号
\usepackage{amsmath}
\usepackage{bm}
\usepackage{array}   
\usepackage{makecell}
\usepackage{multirow}
\usepackage{pifont}
\begin{document}

\title{\paperTitle}

%%%%%%%%% AUTHORS - PLEASE UPDATE
% \and

\author{Jiaqi Li\textsuperscript{1,2}\footnotemark[1] \footnotemark[2] \quad Junshu Tang\textsuperscript{1}\footnotemark[1] \quad  Zhiyong Xu\textsuperscript{1} \quad Longhuang Wu\textsuperscript{1} \\ \quad  Yuan Zhou\textsuperscript{1} \quad Shuai Shao\textsuperscript{1} \quad Tianbao Yu\textsuperscript{1} \quad Zhiguo Cao\textsuperscript{2} \quad Qinglin Lu\textsuperscript{1}\footnotemark[3]\\
\\
\textsuperscript{1} Tencent Hunyuan  \quad \textsuperscript{2} Huazhong University of Science and Technology
 \\
\url{https://hunyuan-gamecraft.github.io/}
}
% \tt\small \{monsantplusone, djhuang\}@shu.edu.cn, tangjs@sjtu.edu.cn, 22210980090@m.fudan.edu.cn, \\ \tt\small186368@zju.edu.cn}

\twocolumn[{
\renewcommand\twocolumn[1][]{#1}
\maketitle
\centering
\vspace{-7mm}
 \includegraphics[width=\linewidth]{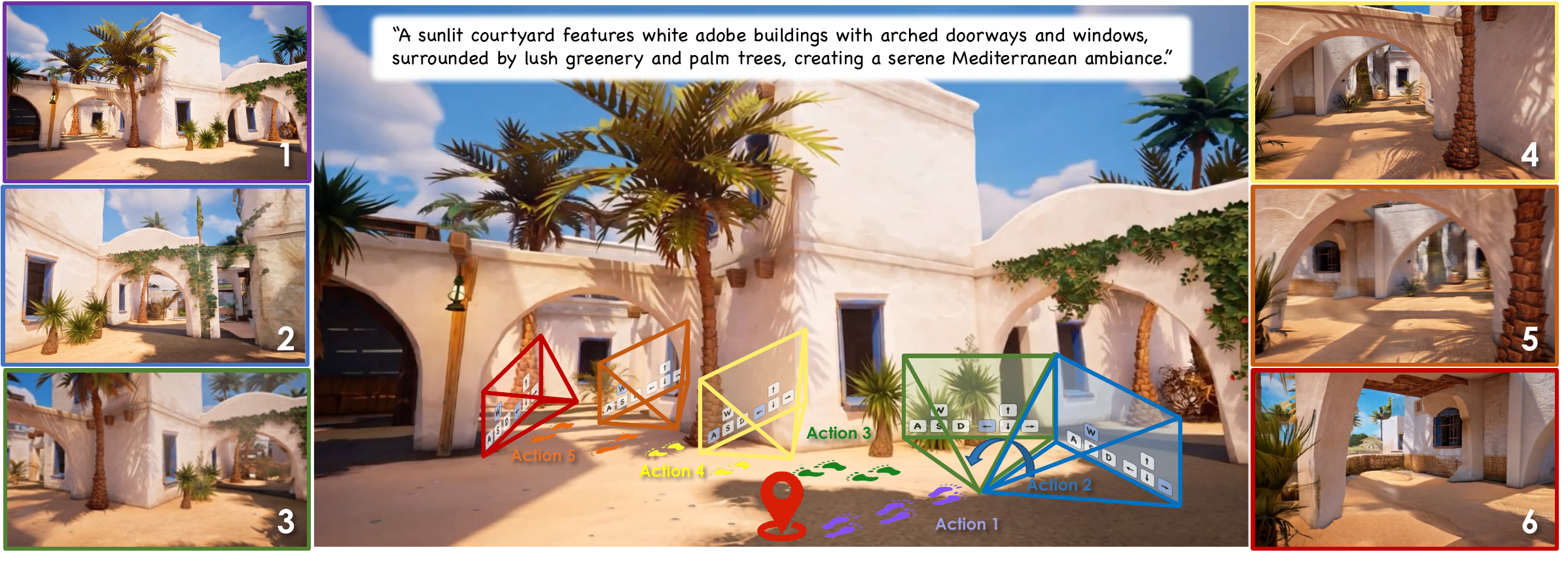}
 
 \captionsetup{type=figure}
\vspace{-3mm}
\caption{\emph{Hunyuan-GameCraft} can create high-dynamic interactive game video content from a single image and corresponding prompt. We simulate a series of action signals. The left and right frames depict key moments from game video sequences generated in response to different inputs. Hunyuan-GameCraft can accurately produce content aligned with each interaction, supports long-term video generation with temporal and 3D consistency, and effectively preserves historical scene information throughout the sequence. In this case, W, A, S, D represent transition movement and ↑, ←, ↓, → denote changes in view angles.}
\vspace{5mm}
\label{fig:teaser}
}]

\renewcommand{\thefootnote}{\fnsymbol{footnote}}
\footnotetext[1]{Equal Contribution.}
\footnotetext[2]{Work is done during the internship at Tencent Hunyuan.}
\footnotetext[3]{Corresponding author.}

\begin{abstract}
\vspace{-3mm}
Recent advances in diffusion-based and controllable video generation have enabled high-quality and temporally coherent video synthesis, laying the groundwork for immersive interactive gaming experiences. However, current methods face limitations in dynamics, generality, long-term consistency, and efficiency, which limit the ability to create various gameplay videos. To address these gaps, we introduce \textbf{Hunyuan-GameCraft}, a novel framework for high-dynamic interactive video generation in game environments. 
To achieve fine-grained action control, we unify standard keyboard and mouse inputs into a shared camera representation space, facilitating smooth interpolation between various camera and movement operations.
Then we propose a hybrid history-conditioned training strategy that extends video sequences autoregressively while preserving game scene information. Additionally, to enhance inference efficiency and playability, we achieve model distillation to reduce computational overhead while maintaining consistency across long temporal sequences, making it suitable for real-time deployment in complex interactive environments. 
The model is trained on a large-scale dataset comprising over one million gameplay recordings across over 100 AAA games, ensuring broad coverage and diversity, then fine-tuned on a carefully annotated synthetic dataset to enhance precision and control. The curated game scene data significantly improves the visual fidelity, realism and action controllability.  
Extensive experiments demonstrate that Hunyuan-GameCraft significantly outperforms existing models, advancing the realism and playability of interactive game video generation.

\begin{figure*}[htbp]
    \centering
    \includegraphics[width=\linewidth]{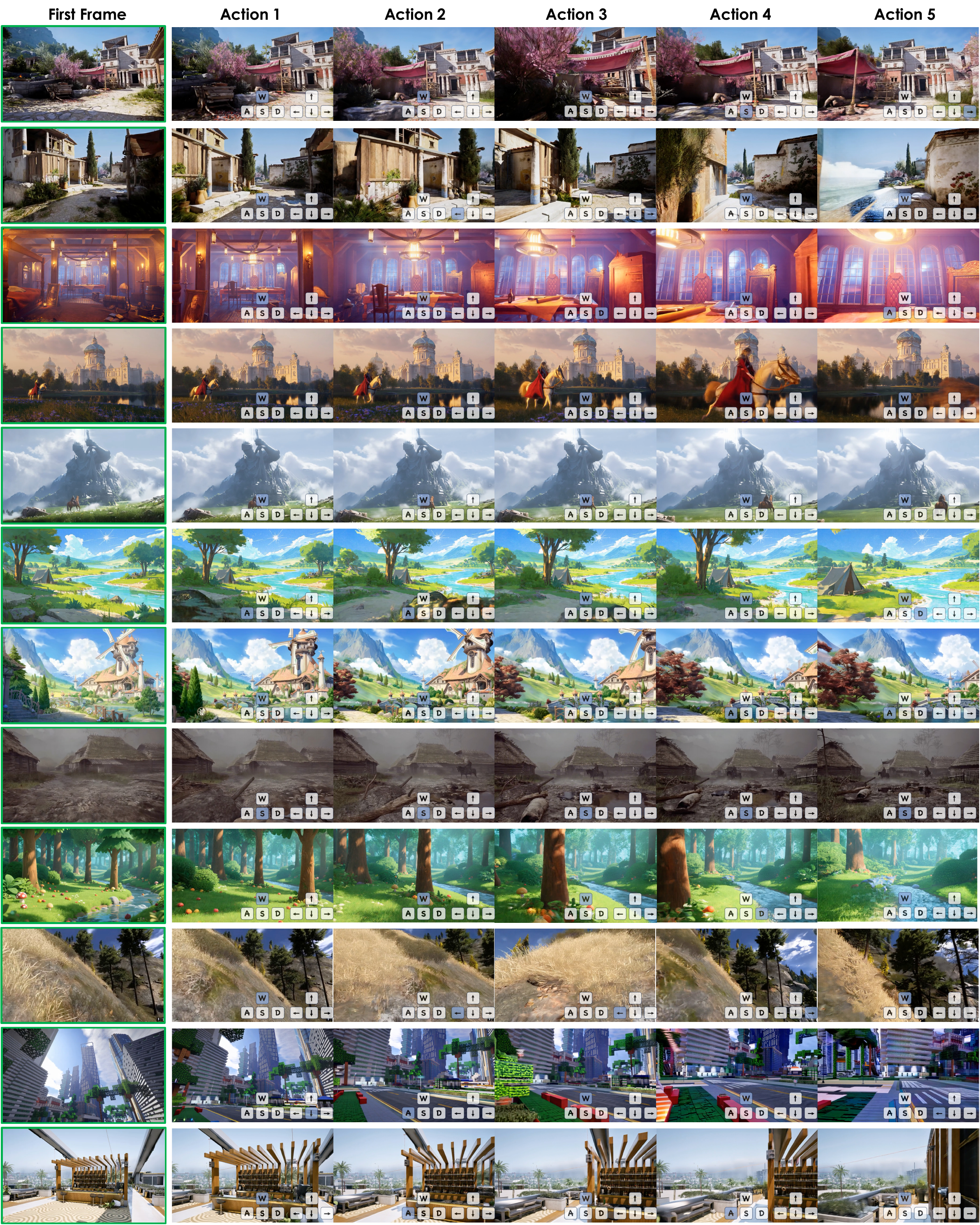}
    \caption{Additional results by \textit{Hunyuan-GameCraft} on multi-actions control. In our case, blue-lit keys indicate key presses. W, A, S, D represent transition movement and ↑, ←, ↓, → denote changes in view angles.}
    \label{fig:figure1}
\end{figure*}

\end{abstract}

\section{Introduction}

The rapid progress in generative modeling has transformed numerous fields, including entertainment and education, and beyond, fueling growing interest in high-dynamic, immersive, generative gaming experiences. Recent breakthroughs in diffusion-based video generation~\cite{blattmann2023stable,blattmann2023align,wang2024motionctrl,chen2024diffusion,li2025hunyuan} have significantly advanced dynamic content creation, enabling high-quality, temporally coherent video synthesis. Moreover, advances in controllable video generation have introduced novel creative forms of dynamic, user-driven video production, expanding the boundaries of interactive digital experiences.

% developing World Models—frameworks capable of synthesizing diverse, interactive, and temporally coherent virtual environments. These models aim to go beyond static scenes or conventional video generation by incorporating interactivity, physics reasoning, and behavioral dynamics, thereby enabling agents or users to engage with simulated 3D/4D environments in a manner that mirrors real-world complexity. World Models serve as foundational tools for applications in simulation, robotics, and next-generation gaming experiences.

\begin{table*}[!ht]\footnotesize
    \centering
    \renewcommand{\arraystretch}{1.5} % 调整行高
    \newcommand{\gou}{{\ding{52}}}
    \newcommand{\cha}{{\ding{55}}}
    \resizebox{0.95\linewidth}{!}{
    \begin{tabular}{c|c|c|c|c|c|c|c|c}
        \toprule
        &\textbf{GameNGen}~\cite{valevski2024diffusion} & \textbf{GameGenX}~\cite{che2024gamegen}  & \textbf{Oasis}~\cite{oasis} & \textbf{Matrix}~\cite{feng2024matrix} & \textbf{Genie 2}~\cite{parkerholder2024genie2} & \textbf{GameFactory}~\cite{yu2025gamefactory} & \textbf{Matrix-Game}~\cite{zhang2025matrixgame} & \textbf{Hunyuan-GameCraft} \\
        \hline
        % Release Time & ICLR 2025 & ICLR 2025 & 2024.10.31 & 2024.12.4 & 2024.12.4 & - & - & - \\
        % \hline
        Game Sources & DOOM & AAA Games& Minecraft & AAA Games & Unknown & Minecraft & Minecraft & AAA Games\\
        \hline
        Resolution & $240$p & $720$p & $640\times360$ & $720$p & $720$p & $640\times360$ & $720$p & $720$p\\
        \hline
        % Control Granularity & Frame-level & Video-level & Frame-level & Frame-level & Frame-level & Frame-level & Clip-level & Frame-level\\
        % \hline
        % Technical Paper & \gou & \gou & \cha & \gou & \cha & \gou & \gou & \gou \\
        % \hline
        %  & - & - & - & - & - & - & - & - \\
        % \hline
        % Testable Model & \cha & \cha & \cha & \cha & \cha & - & - & - \\
        % \hline
        % Available Dataset & \cha & \gou & \cha & \cha & \cha & \gou & - & - \\
        % \hline
        Action Space & Key & Instruction & Key + Mouse & 4 Keys & Key+Mouse & 7 Keys+Mouse & 7 Keys+Mouse & Continous \\
        \hline
        Scene Generalizable & \cha & \cha & \cha & \gou & \gou & \gou & \gou & \gou \\
        \hline
        Scene Dynamic & \gou & \gou & \cha & \gou & \cha & \gou & \cha & \gou \\
        \hline
        Scene Memory & \cha & \cha & \cha & \cha & \cha & \cha & \gou & \gou \\
        \bottomrule
    \end{tabular}}
    \caption{Comparison with recent interactive game models. Hunyuan-GameCraft serves as a model capable of generating infinitely long game videos conditioned on continuous action signals, while maintaining strong generalization, high temporal dynamics, and effective preservation of historical scene information.}
    \label{tab:compare_works}
\end{table*}

Recent advances in visual generation have explored spatial intelligence, the analysis and creation of coherent spatial scenes. These models focus on interactivity and exploration, enabling dynamic 3D/4D environments with spatiotemporal coherence. For example, WorldLabs~\cite{worldlabs2024} demonstrates the potential for reconstructing high-fidelity 3D environments from static imagery, while Genie 2~\cite{parkerholder2024genie2} introduces latent action modeling to enable physics-consistent interactions over time.
Despite these advances, current approaches still struggle with significant limitations in critical areas such as real-time dynamic scene element fidelity, long-sequence consistency, and computational efficiency, limiting their applicability in high-dynamic, playable interactive scenarios. Notably, in game interaction modeling, real-time interactive generation and high dynamicity constitute fundamental components of player experience.

To address these challenges, we introduce Hunyuan-GameCraft, a novel framework designed for high-dynamic, action-controllable video synthesis in game environments. Built upon a text-to-video foundation model, HunyuanVideo~\cite{kong2024hunyuanvideo}, our method enables the generation of temporally coherent and visually rich gameplay footage conditioned on discrete user actions. 
We unify a broad set of standard keyboard and mouse inputs (e.g., W, A, S, D, arrow keys, Space) into a shared camera representation space, which unified embedding supports smooth interpolation between various camera and movement operations, ensuring physical plausibility while enabling cinematic flexibility in user-driven interactions, for example, speeding up.

To maintain long-term consistency in interactive game video generation, prior works~\cite{chen2024diffusion,henschel2024streamingt2v,lu2024freelong} have primarily focused on training-free extensions, streaming denoising or last-frame conditioning. However, these approaches often suffer from quality degradation and temporal inconsistency with causal VAEs~\cite{yang2021causalvae}.
We propose a novel hybrid history-conditioned training strategy that autoregressively extends sequences while preserving scene information, using historical context integration and a mask indicator to address error accumulation in autoregressive generation. Moreover, to improve inference efficiency and playability, we implement the model distillation acceleration strategy~\cite{wang2024phased}, which reduces computational overhead while maintaining consistency across long temporal sequences, making our framework suitable for real-time deployment in complex interactive environments.
% Specifically, we integrate multiple chunks of historical context, including previous last frame, and past denoised sequences. We define a mask indicator to distinguish the condition and the denoising part.
% This allows for realistic transitions and sustained narrative flow, addressing the common issue of error accumulation or temporal incoherence in autoregressive generation.Moreover, to improve inference efficiency and playability, we implement the model distillation acceleration strategy~\cite{wang2024phased}, which reduces computational overhead while maintaining consistency across long temporal sequences, making our framework suitable for real-time deployment in complex interactive environments.

% By bridging the gap between high-quality video generation and real-time user interaction, our proposed GameCraft provides a scalable and practical foundation for the future of intelligent, interactive content creation in games and virtual simulations.
We evaluate our Hunyuan-GameCraft on both curated game scenes and general styles, obtaining a significant lead over current models. In summary, our contributions are:
\begin{itemize}[leftmargin=15pt]
  \item We propose Hunyuan-GameCraft, a novel interactive game video synthesis framework for dynamic content creation in game scenes, enabling users to produce content through customized action input. 
  \item We unify the discrete keyboard/mouse action signals into a shared continuous action space, supporting more complex and fine-grained interactive inputs, such as speed, angle, etc.
  \item We introduce a novel hybrid history-condition training strategy that maintains long-term spatial and temporal coherency across various action signals.
  \item We implement model distillation to speed up the inference speed which improves the interaction experience. 
\end{itemize}
% Extensive experiments show that GameCraft significantly outperforms the state-of-the-art interactive game video generation methods in terms of image quality, dynamics, interaction accuracy, and long sequence consistency.
% \end{itemize}
\section{Related Work}

\subsection{Interactive Game Scene World Model}
Recent research has gradually focused on incorporating video generation models to enhance dynamic prediction and interaction capabilities in game scenes. We conduct a survey on recent works, as shown in Tab.~\ref{tab:compare_works}.
WorldDreamer~\cite{wang2024worlddreamer} proposes constructing a general world model by predicting masked tokens, which supports multi-modal interaction and is applicable to natural scenes and driving environments. GameGen-X~\cite{che2024gamegen}, a diffusion Transformer model for open-world games, integrates multi-modal control signals to enable interactive video generation. The Genie series~\cite{parkerholder2024genie2} generates 3D worlds from single-image prompts, while the Matrix model leverages game data with a streaming generation format to infinitely produce content through user actions.
\subsection{Camera-Controlled Video Generation}
Motionctrl~\cite{wang2024motionctrl} uses a unified and flexible motion controller designed for video generation, which independently controls the movement of video cameras and objects to achieve precise control over the motion perspectives in generated videos. CameraCtrl~\cite{he2024cameractrl} employs Plücker embedding as the primary representation for camera parameters, training only the camera encoder and linear layers to achieve camera control. Furthermore, the recent approach CameraCtrl II~\cite{he2025cameractrl} constructs a high-dynamics dataset with camera parameter annotations for training, and designs a lightweight camera injection module and training scheme to preserve the dynamics of pretrained models. 
% By harnessing the generative capabilities of pre-trained text-to-video models, ReCamMaster~\cite{bai2025recammaster} employs a simple yet effective video conditioning mechanism that fuses the temporal features of input videos with camera parameters.
\subsection{Long Video Extension}

Generating long videos poses challenges in maintaining temporal consistency and high visual quality over extended durations. 
Early methods used GAN to explore long video generation~\citep{skorokhodov2022stylegan}.
% However, GAN training is very unstable and the quality of generation is limited. 
With the popularity of diffusion, some methods began to try to solve the problem using diffusion model.
% LVDM~\citep{he2022latent} leverages a low-dimensional 3D latent space with hierarchical diffusion to generate videos exceeding one thousand frames. 
% Lavie~\citep{wang2025lavie} improves quality through cascaded latent diffusion with rotary positional encoding and joint image-video fine-tuning. 
StreamingT2V~\citep{henschel2024streamingt2v} introduces short-term and long-term memory blocks with randomized blending to ensure consistency and scalability in text-to-video generation. 
% NUWA-XL~\citep{yin2023nuwa} adopts a coarse-to-fine framework: a global model generates keyframes, while local models fill in the in-between frames for better structure and detail. 
% Seine~\citep{chen2023seine} proposes a short-to-long diffusion model using random masking to create smooth transitions guided by text. Meanwhile, FreeNoise~\citep{qiu2023freenoise} offers a training-free method using noise rescheduling, allowing pretrained models to extend to longer sequences without additional training. 
In addition, some methods also explore different paradigms, such as next frame prediction~\citep{gao2024vid,gu2025long}, combining next-token and full-sequence diffusion (DiffusionForcing)~\citep{chen2024diffusion} and test-time training ~\citep{dalal2025one}. Compared with previous methods, we propose a novel hybrid history-conditioned training strategy that extends video sequences in an autoregressive way while effectively preserving game scene information, under a diffusion paradigm.

\section{Dataset Construction}
\label{sec:data}
%Our Dataset Construction work comprises three key components, with full implementation details provided in supplementary materials.
\subsection{Game Scene Data Curation}
% To facilitate the generation of high-fidelity and high-dynamic gameplay videos, we curated a large and diverse dataset of gameplay recordings from over 100 AAA first-person perspective titles, including \textit{Assassin's Creed}, \textit{Red Dead Redemption}, \textit{Hogwarts Legacy}, \textit{Cyberpunk 2077}, and others. These recordings were carefully selected for interactive video generation to encompass a broad range of environments, actions, and visual styles, providing rich coverage of high-resolution graphics, dynamic lighting, diverse settings, and complex in-game interactions. 

\begin{figure}
    \centering
    \includegraphics[width=\linewidth]{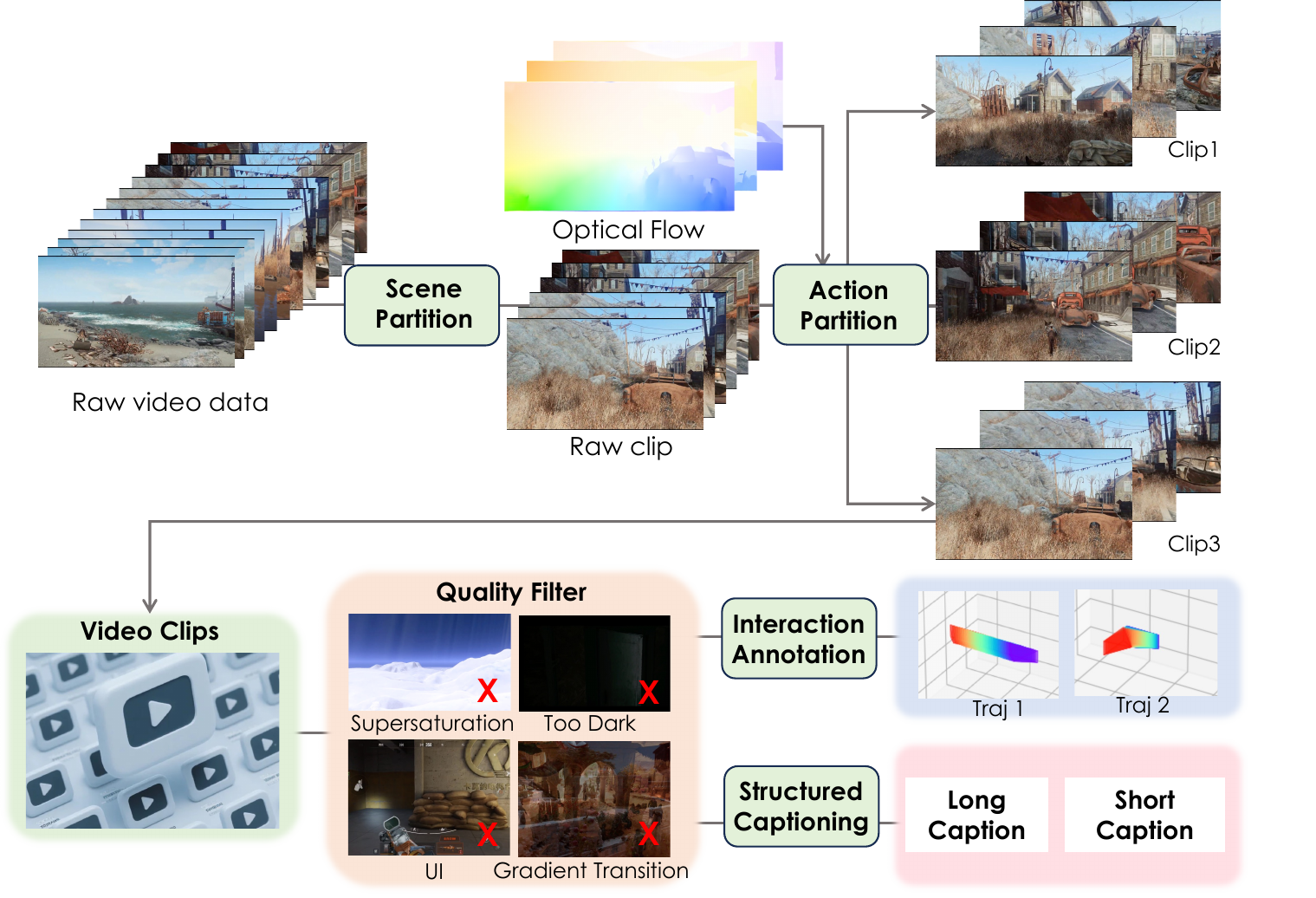}
    \caption{Dataset Construction Pipeline. It consists of four pre-processing steps:
Scene and Action-aware Data Partition, Data Filtering, Interaction Annotation and structured captioning.}
    \label{fig:data}
\end{figure}

We curate over 100 AAA titles, such as \textit{Assassin's Creed}, \textit{Red Dead Redemption}, and \textit{Cyberpunk 2077}, to create a diverse dataset with high-resolution graphics and complex interactions.
As shown in Fig~\ref{fig:data}, our end-to-end data processing framework comprises four stages that addresses annotated gameplay data scarcity while establishing new standards for camera-controlled video generation.
%to ensure GameEngine-like quality and accurate interactive outputs for gameplay video generation.

\begin{figure*}[htb!]
    \centering
    \includegraphics[width=\linewidth]{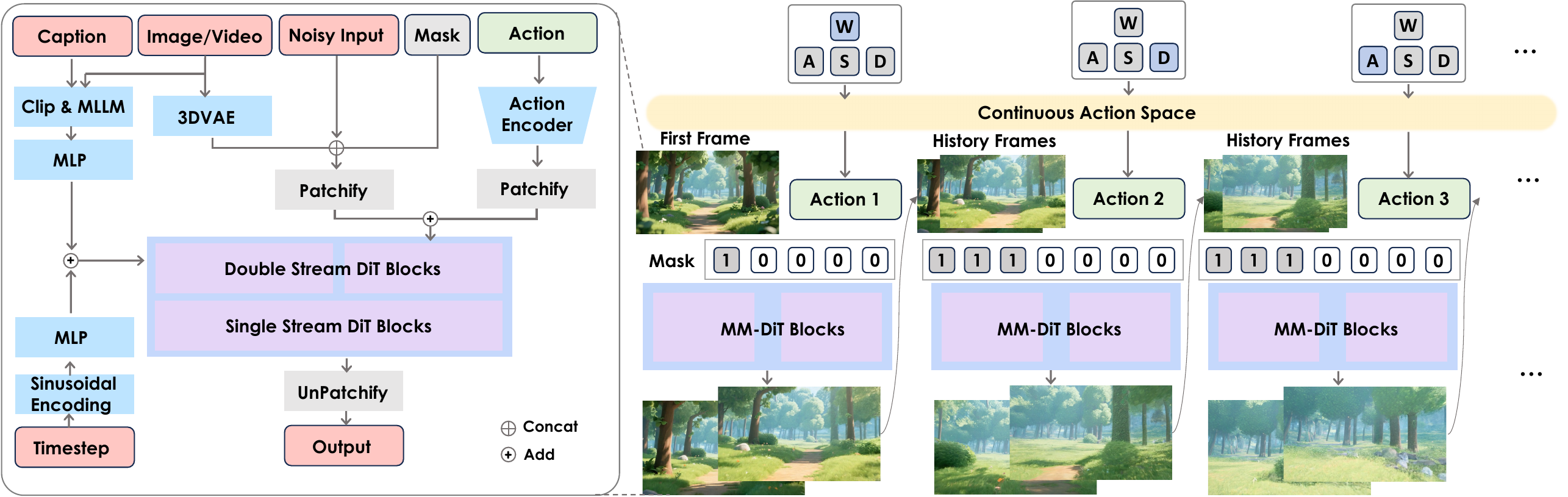}
    \caption{Overall architecture of Hunyuan-GameCraft. Given a reference image and the corresponding prompt, the keyboard or mouse signal, we transform these options to the continuous camera space. Then we design a light-weight action encoder to encode the input camera trajectory. The action and image features are added after patchify. For long video extension, we design a variable mask indicator, where 1 and 0 indicate history frames and predicted frames, respectively. }
    \label{fig:gamecraft}
\end{figure*}

\noindent \textbf{Scene and Action-aware Data Partition.}
%We propose a multi-stage video data partition mechanism, including a scene-level and an action-level video segmentation. Specifically, given a 2-3 hour long gameplay recording, we first leverage PySceneDetect~\cite{} to split the raw video into environment and semantic coherent single-shot clips, in our case, 6-second segments. After the partition, we collect over 1,000,000 raw gameplay clips at 1080p resolution. To ensure temporal consistency and minimize artifacts from external input variations, we then compute the dense optical flow in each clip using RAFT~\cite{}. We then calculate the frame-level gradients of the optical flow and identify frames where the gradient exceeds a defined threshold as action segmentation points. This allows us to detect sudden camera or player movements (e.g., rapid aiming or scene transitions) as candidate split points. This hierarchical partitioning mechanism facilitates stronger alignment between instructional input and video content, thereby reducing the complexity of subsequent interactive video generation training.
We introduce a two-tier video partitioning approach (scene-level and action-level). Using PySceneDetect~\cite{Castellano_PySceneDetect}, we segment 2-3 hour gameplay recordings into 6-second coherent clips (1M+ clips at 1080p). RAFT~\cite{teed2020raft} computes optical flow gradients to detect action boundaries (e.g., rapid aiming), enabling precise alignment for video generation training.

\noindent \textbf{Data Filtering.}
%To enhance the fidelity and motion intensity of video synthesis, we leverage various filters to help to filter data from different perspectives. 
%Firstly, we employ a quality assessment operator~\cite{} to clarify the visual quality of video clips and then remove low-fidelity segments. Additionally, we use OpenCV~\cite{} to compute the average luminance of each clip and discard scenes that are nearly completely black. 
%Besides, we also observed that some videos have gradient transitions which PySceneDetect cannot detect, so we leverage a VLM~\cite{} to detect the gradient effect for videos. 
To enhance synthesis quality, we employ quality assessment~\cite{kolors} to remove low-fidelity clips, apply OpenCV~\cite{bradski2000opencv}-based luminance filtering to eliminate dark scenes, and utilize VLM~\cite{wang2024qwen2}-based gradient detection for comprehensive data filtering from multiple perspectives.

% Flag abrupt scene transitions using optical flow variance thresholds for later stabilization.
% \textbf{Action Segmentation and Analysis:} 
% To temporally partition long gameplay sequences into distinct action units (e.g., turn left, move forward), we compute dense optical flow from RAFT~\cite{} to compute dense optical flow and leverage gradients as motion saliency indicators. According to the flow vectors, sudden movements in camera or player motion (e.g., rapid aiming, scene transitions) are detected as candidate split points.
% After partition, we reconstruct 6-DoF camera trajectories via Monst3r~\cite{} to model viewpoint dynamics.
% As for game video data annotation, we follow the design introduced in Section~\ref{}, focusing on the description of camera motion and various environments.  

\noindent \textbf{Interaction Annotation.}
%After partitioning and filtering the gameplay recordings, we reconstruct six-degree-of-freedom (6-DoF) camera trajectories using Monst3R~\cite{}, a state-of-the-art monocular camera pose estimation method. This process enables accurate modeling of the camera’s viewpoint dynamics, including both translational and rotational motion across time. For each resulting video clip, we annotate it with its corresponding 6-DoF camera trajectory, providing frame-by-frame information on the camera's position and orientation. These precise trajectory annotations are critical for the interaction modeling in the training of video generation.
We reconstruct 6-DoF camera trajectories using Monst3R~\cite{zhang2024monst3r} to model viewpoint dynamics (translational/rotational motion). Each clip is annotated with frame-by-frame position/orientation data, which is essential for video generation training.

\noindent \textbf{Structured Captioning.}
% In order to fully utilize the unlabeled live game data for interaction model training, inspired by previous work\cite{Anonymous:1976:POT}, we obtain the largest current dataset designed for interactive game video generation from the web, which achieves the richest diversity and the most accurate labeling currently available through our proposed data cleaning and labeling pipeline.
%For video captioning, inspired by prior works~\cite{}, we employ a hierarchical multi-length caption strategy. We leverage VLM designed for game video data~\cite{} and generate short and long captions. The short caption includes approximately 30 characters, briefly summarizing the main content of the video. And the long description includes more than 100 characters, comprehensively elaborating on the video detail. During model training, the captions were randomly sampled and dropped.
For video captioning, we implement a hierarchical strategy using game-specific VLMs~\cite{wang2024qwen2} to generate: 1) concise 30-character summaries and 2) detailed 100+ character descriptions. These captions are randomly sampled during training.

\subsection{Synthetic Data Construction}
%To further enhance the model's capability in handling precise camera motion and interactive signals, we rendered almost 3000 high-quality motion sequences from curated 3D scene assets for fine-tuning purposes. Each scene is systematically sampled with multiple starting positions, generating camera trajectories that encompass isolated translations, pure rotations, and their composite motions. To maximize data diversity, all trajectories are re-rendered at varying speeds to scale up samples.

%Our experiments demonstrate that strategically introducing these high-precision, fully controllable rendered sequences during later stages of multi-phase training yields significant improvements. Specifically, this approach enhances the accuracy in predicting both motion direction and velocity, while simultaneously improving the temporal coherence during large-scale viewpoint transitions. The synthetic data proves particularly valuable for establishing robust geometric priors and maintaining consistency in complex camera movements that are challenging to learn from, coupled with real-world samples.
We rendered about 3,000 high-quality motion sequences from curated 3D assets, systematically sampling multiple starting positions to generate diverse camera trajectories (translations, rotations, and composites) re-rendered at varying speeds. Our multi-phase training strategy demonstrates that introducing high-precision rendered sequences significantly improves motion prediction accuracy and temporal coherence during viewpoint transitions, while establishing essential geometric priors for complex camera movements that complement real-world samples.

\subsection{Distribution Balancing Strategy}
%Building upon the hybrid training framework, combining both datasets, we identified an inherent bias in the data distribution - the majority of camera trajectories in both web-crawled gameplay footage and reconstructed sequences predominantly feature forward-moving motions. To enable the generative model to uniformly learn control signals across all possible directions, we implemented a comprehensive distribution balancing strategy. In addition to fine-tuning with more uniformly distributed rendered data during later training stages, we introduced two key augmentation techniques: stratified sampling based on camera trajectory start-end vectors and temporal inversion augmentation.

%The stratified sampling approach explicitly balances the training batches by ensuring proportional representation of trajectories across different directional sectors in 3D space. Meanwhile, the temporal inversion augmentation creates additional training samples by reversing existing trajectories, effectively doubling the coverage of backward motions without requiring additional data collection. These techniques collectively enhance the model's interactive controllability while significantly improving training stability, as evidenced by more consistent performance across all motion directions during evaluation.
Leveraging a hybrid training framework with combined datasets, we addressed inherent forward-motion bias in camera trajectories via a two-pronged strategy: 1) stratified sampling of start-end vectors to balance directional representation in 3D space and 2) temporal inversion augmentation to double backward motion coverage. Combined with late-stage fine-tuning using uniformly distributed rendered data, these techniques enhanced control signal generalization, training stability, and cross-directional performance consistency.

\section{Method}

In this paper, we propose Hunyuan-GameCraft, a high-dynamic interactive game video generation model based on a previously open-sourced MM-DiT~\cite{esser2024scaling} based text-to-video model, HunyuanVideo~\cite{kong2024hunyuanvideo}. The overall framework is shown in Fig~\ref{fig:gamecraft}. 
To achieve fine-grained controllable game video synthesis with temporal coherence, we first unify diverse common keyboard/mouse options in games (W, A, S, D, ↑, ←, ↓, →, Space, etc.) into a shared camera representation space (Sec.~\ref{sec:action}) and design a light-weight action encoder to encode the camera trajectory(Sec.~\ref{sec:injection}).
Then, we propose a hybrid history-conditioned video extension approach that autoregressively denoise new noisy latent conditioned on historical denoised chunks (Sec.~\ref{seb:long}).
Finally, to accelerate the inference speed and improve the interaction experience, we implement the model distillation, based on Phased Consistency Model~\cite{wang2024phased}. This distillation achieves a 10–20× acceleration in inference speed, reducing latency to less than 5s per action (Sec.~\ref{sec:fast}).
% Compared with previous interactive game video generation works~\cite{} or world models, the foundation transformer-based model brings strong generalization capabilities and helps to generate highly dynamic, realistic videos for interactive video generation. 

\subsection{Continuous Action Space and Injection}
\label{sec:action}
To achieve fine-grained control over the generated content for enhanced interactive effects, we define a subset action space $\mathcal{A}$ within the camera parameter $\mathcal{C} \subseteq \mathbb{R}^n$ dedicated to continuous and intuitive motion control injection:
\begin{equation}
\mathcal{A} \coloneqqb \Bigg\{ \mathbf{a} = \Big( \mathbf{d}_{\text{trans}}, \mathbf{d}_{\text{rot}}, \alpha, \beta \Big) \;\Bigg|\; 
\begin{aligned} 
&\mathbf{d}_{\text{trans}} \in \mathbb{S}^2, \quad \mathbf{d}_{\text{rot}} \in \mathbb{S}^2, \\ 
&\alpha \in [0, v_{\text{max}}], \quad \beta \in [0, \omega_{\text{max}}] 
\end{aligned} \Bigg\}\!.
\end{equation}
$\mathbf{d}_{\text{trans}}$ and $\mathbf{d}_{\text{rot}}$ are unit vectors defining the translation and rotation direction on the 2-sphere space $\mathbb{S}^2$, respectively. Scalars $\alpha$ and $\beta$ are used for controlling translation and rotation speed, bounded by maximum velocity $v_{\text{max}}$ and $\omega_{\text{max}}$. Specifically, they are the differential modulus of relative velocity and angle during frame-by-frame motion.

Building upon prior knowledge of gaming scenarios and general camera control conventions, we eliminate the degree of freedom in the roll dimension while incorporating velocity control. This design enables fine-grained trajectory manipulation that aligns with user input habits. Furthermore, this representation can be seamlessly converted into standard camera trajectory parameters and Plücker embeddings.
% \subsection{Action Control Injection}
\label{sec:injection}
% We systematically investigate multiple interactive signals injection approaches within the MMDiT full attention architecture. Specifically, as illustrated in Fig.~\ref{fig:gamecraft}, 
% we experiment with 
% three distinct injection approaches, including (i)~Token Addition, (ii)~Token Concatenation, (iii)~Channel-wise Concatenation. 
% After balancing parameter efficiency and generation quality, 
% As for camera information injection, multiple previous arts 
Similar with previous camera-controlled video generation arts, we design a light-weight camera information encoding network that aligns Plücker embeddings with video latents. Unlike previous approaches that employ cascaded residual blocks or transformer blocks to construct Plücker embedding encoders, our encoding network consists solely of a limited number of convolutional layers for spatial downsampling and pooling layers for temporal downsampling. A learnable scaling coefficient is incorporated to automatically optimize the relative weighting during token-wise addition, ensuring stable and adaptive feature fusion.

Then we adopted the token addition strategy to inject camera pose control into the MM-DiT backbone. Dual lightweight learnable tokenizers are used to achieve efficient feature fusion between video and action tokens, enabling effective interactive control. Additional ablation studies and comparative analyses are detailed in Sec.~\ref{sec:ab}.

% Furthermore, we introduce an effective camera pose encoding network that aligns Plücker embeddings with video latents. Unlike previous approaches that employ cascaded residual blocks or transformer blocks to construct Plücker embedding encoders~\cite{}, our encoding network consists solely of a limited number of convolutional layers for spatial downsampling and pooling layers for temporal downsampling. A learnable scaling coefficient is incorporated to automatically optimize the relative weighting during token-wise addition, ensuring stable and adaptive feature fusion.

Leveraging the robust multimodal fusion and interaction capabilities of MM-DiT backbone, our method achieves state-of-the-art interactive performance despite significant encoder parameter reduction, while maintaining negligible additional computational overhead.

\begin{figure}[t]
    \centering
    \includegraphics[width=\linewidth]{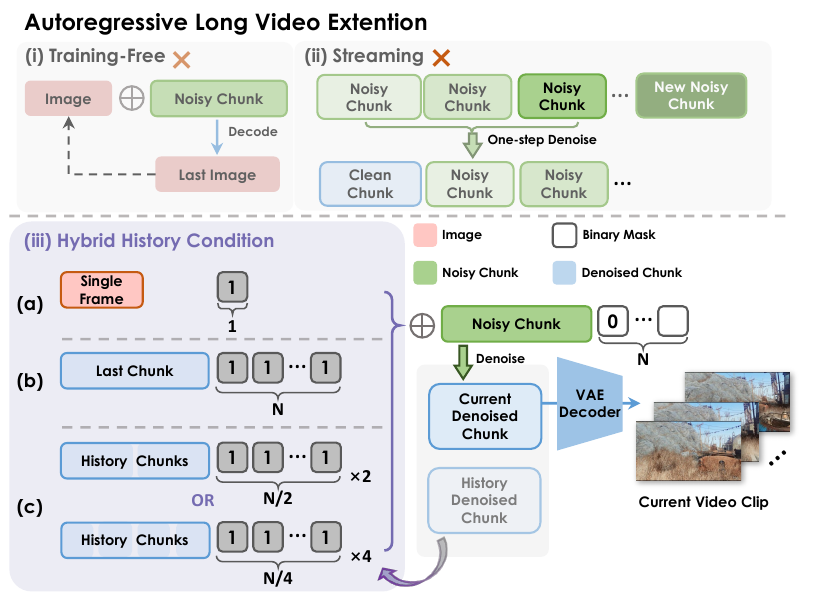}
    \caption{Comparison of different autoregressive long video extension schemes. (i) Training-free inference. (ii) Streaming generation. (iii) Hybrid history condition proposed in this paper.}
    \label{fig:longinfer}
\end{figure}

\subsection{Hybrid history conditioned Long Video Extension}
\label{seb:long}
\begin{figure}[t]
    \centering
    \includegraphics[width=\linewidth]{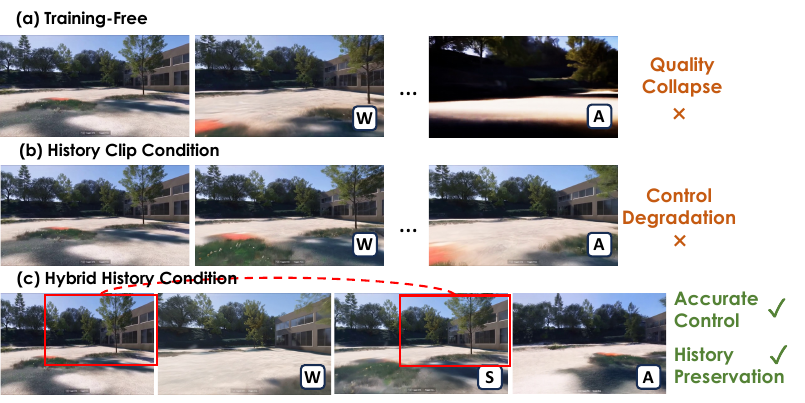}
    \caption{Analysis on different video extension schemes. Baseline (a) is a naive solution using training-free inference from single images, and it will lead to obvious quality collapse. Using history clip condition (b) will result in control degradation. With our proposed hybrid history condition (c), the model can achieve accurate action control and history preservation (see {\color{red}red} box). W, A, S denote moving forward, left and backward. }
    \label{fig:ab_long}
\end{figure}

Consistently generating long or potentially infinite-length videos remains a fundamental challenge in interactive video generation. As shown in Fig~\ref{fig:longinfer}, current video extrapolation approaches can be categorized into three main paradigms: (1) training-free inference from single images, (2) rolling streaming generation with non-uniform noise windows, and (3) chunk-wise extension using historical segments. 
As shown in Fig~\ref{fig:ab_long}(a), 
% uring our preliminary exploration, we systematically evaluated all three approaches. 
training-free methods lack insufficient historical context during extrapolation, leading to inconsistent generation quality and frequent scene collapse in iterative generation. The streaming approach shows significant architectural incompatibility with our image-to-video foundation model, where the causal VAE's uneven encoding of initial versus subsequent frames fundamentally limits efficiency and scalability. To address these limitations, we investigate hybrid-conditioned autoregressive video extension, where multiple guidance conditions are mixed during training to achieve high consistency, fidelity, and compatibility.

% To address this, we systematically investigated various autoregressive schemes to optimize the trade-off between temporal consistency, generation quality, and computational cost. 
As illustrated in Fig.~\ref{fig:longinfer}, we define each autoregressive step as a chunk latent denoising process guided by head latent and interactive signals. The chunk latent, serving as a global representation by causal VAE, is subsequently decoded into a temporally consistent video segment that precisely corresponds to the input action. Head condition can be different forms, including (i) a single image frame latent, (ii) the final latent from the previous clip, or (iii) a longer latent clip segment. Hunyuan-GameCraft achieves high-fidelity denoising of chunk latents through concatenation at both condition and noise levels. An additional binary mask assigns value 1 to head latent regions and 0 to chunk segments, enabling precise control over the denoising part. Within the noise schedule, the preceding head condition remains noise-free as clean latent, which guides subsequent noisy chunk latents through flow matching to progressively denoise and generate new clean video clips for the next denoising iteration.

% In our framework, each input action maps to an output atom interactive video clip. The chunk latent structure, which serves as a global representation of the atom clip, is determined by our causal video VAE architecture. Through extensive experimentation to balance the critical tradeoffs between interactive latency, temporal consistency, and motion fluidity, we empirically established an optimal chunk latent length of 33 frames.
% The chunk latent structure is determined by our causal video VAE~\cite{} during training, configured as 33 frames in our experiments. 

We conduct extensive experiments on the three aforementioned head conditions, as detailed in Fig~\ref{fig:ab_long}. The results demonstrate that autoregressive video extension shows improved consistency and generation quality when the head condition contains more information, while interactive performance decreases accordingly. This trade-off occurs because the training data comes from segmented long videos, where subsequent clips typically maintain motion continuity with preceding ones. As a result, stronger historical priors naturally couple the predicted next clip with the given history, which limits responsiveness to changed action inputs. However, richer reference information simultaneously enhances temporal coherence and generation fidelity. 
\begin{figure*}[htb!]
    \centering
    \includegraphics[width=\linewidth]{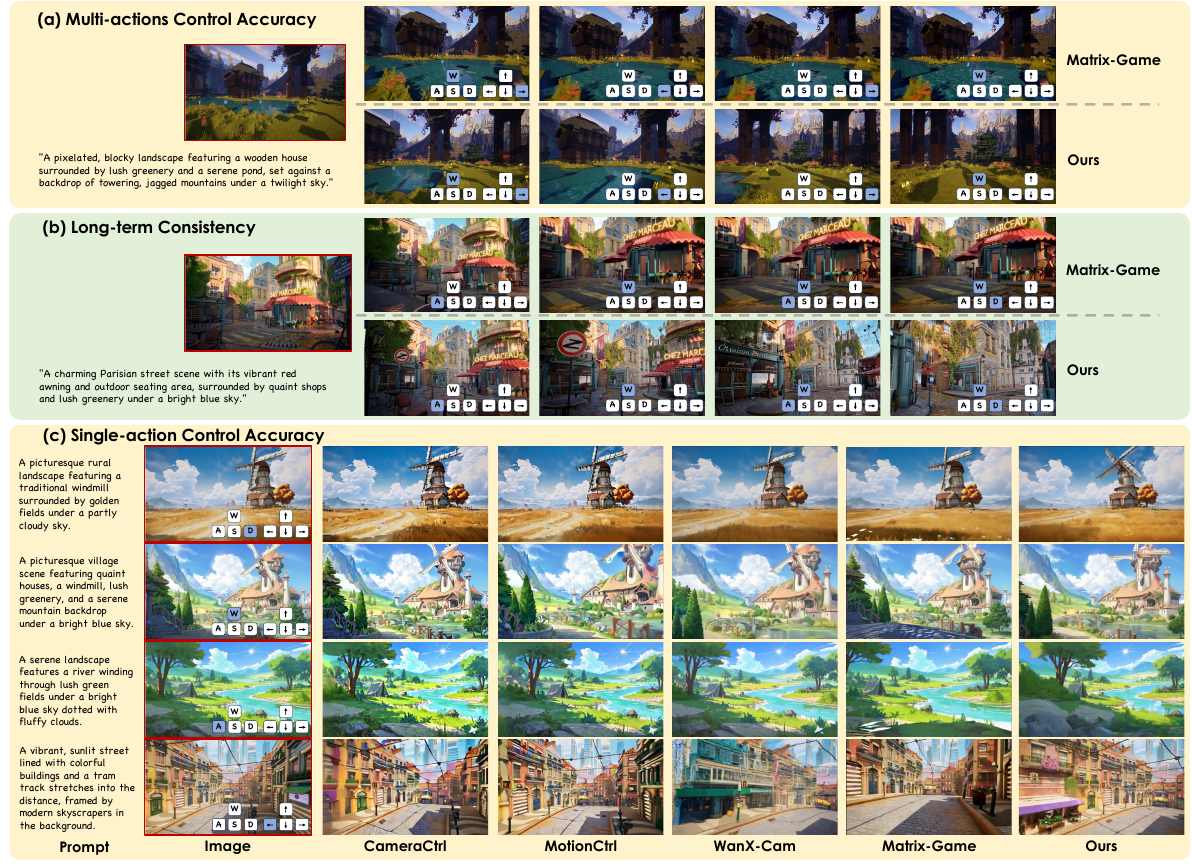}
    \caption{Qualitative comparison on the test benchmark. We compare with Matrix-Game on multi-actions control accuracy and long-term consistency. And we compare with other camera-controlled video generation arts CameraCtrl, MotionCtrl and WanX-Cam on single-action control accuracy. In our case, blue-lit keys indicate key presses. W, A, S, D represent transition movement and ↑, ←, ↓, → denote changes in view angles.  }
    \label{fig:qualitative_compare}
\end{figure*}

To address this trade-off, in addition to constructing training samples
%with abrupt action changes 
and applying stratified sampling, hybrid-conditioned training is proposed to mix all three extension modes during training to jointly optimize both interactive capability and generation consistency. This hybrid approach achieves state-of-the-art performance by reasonably balancing these competing objectives.
The hybrid-conditioned paradigm also provides practical deployment benefits. It successfully integrates two separate tasks (initial frame generation and video extension) into a unified model. This integration enables seamless transitions between generation modes without requiring architectural modifications, making the solution particularly valuable for real-world applications that demand both flexible control and coherent long-term video generation.
% This approach enables stable long-term generation while preserving local details, with the head condition mechanism effectively maintaining inter-clip consistency without requiring expensive memory buffers or complex alignment modules. The binary masking strategy provides a lightweight yet effective solution for conditioning while minimizing computational overhead during the autoregressive process.

\subsection{Accelerated Generative Interaction}
\label{sec:fast}
To enhance the gameplay experience and enable accelerated interaction with the generated game videos, we further extend our approach by integrating acceleration techniques. A promising direction involves combining our core framework with Consistency Models~\cite{luo2023latent}, a state-of-the-art method for accelerating diffusion-based generation. In particular, we adopt the Phased Consistency Model (PCM)~\cite{wang2024phased}, which distills the original diffusion process and classifier-free guidance into a compact eight-step consistency model.
To further reduce computational overhead and improve inference efficiency, we introduce Classifier-Free Guidance Distillation. This approach defines a distillation objective that trains the student model to directly produce guided outputs without relying on external guidance mechanisms, the object function is designed as:
\begin{equation}
\begin{aligned}
    &L_{cfg} = \mathbb{E}_{w\sim p_w, t\sim U[0,1]}[|| \hat{u_\theta}(z_t, t, w, T_s) - u_\theta^{s}(z_t, t, w, T_s)||^{2}_{2}],\\
    &\hat{u_\theta}(z_t, t, w, T_s) = (1+w)u_(z_t, t, T_s) - wu_\theta(z_t, t, \empty)
\end{aligned}
\end{equation}
where $T_s$ denotes the prompt.
Through this integration, we achieve up to a 20× speedup in inference, reaching real-time rendering rates of 6.6 frames per second (FPS), thereby significantly enhancing the interactivity and playability of our system.

\section{Experiment}

\subsection{Experimental Setup}
\noindent \textbf{Implementation Details.}
Hunyuan-GameCraft builds upon text-to-video foundation model HunyuanVideo ~\cite{kong2024hunyuanvideo}, implementing a latent mask mechanism and hybrid history conditioning to achieve image-to-video generation and long video extension. The experiments employ full-parameter training on 192 NVIDIA H20 GPUs, conducted in two phases with a batch size of 48. The first phase trains the model for 30k iterations at a learning rate of $3\times10^{-5}$ using all collected game data and synthetic data at their original proportions. 
% This approach preserves the foundation model's dynamic performance while facilitating rapid adaptation to action signal injection. 
The second phase introduces data augmentation techniques, as described in Sec.~\ref{sec:data}, to balance action distributions, while reducing the learning rate to $1\times10^{-5}$ for an additional 20,000 iterations to enhance generation quality and interactive performance. The hybrid history condition maintains specific ratios: $0.7$ for single historical clip, $0.05$ for multiple historical clips, and $0.25$ for single frame. The system operates at $25$ fps, with each video chunk comprising $33$-frame clips at 720p resolution. 
% This configuration enables effective segmentation of weakly action-coupled sequences in training data, thereby improving motion fluidity and output video quality.

% \noindent \textbf{Training Receipt}

\noindent \textbf{Evaluation Datasets.}
% Our test set comprises 75 all-kind images collected from online sources, covering various gaming scenarios. This composition enables comprehensive comparisons between game scene interaction models and real-world camera control generation approaches. To further evaluate generalization, we compile a diverse qualitative assessment set comprising aesthetically curated images and videos across synthetic and natural domains. 
We curate a test set of 150 diverse images and 12 different action signals, sourced from online repositories, spanning gaming scenarios, stylized artwork, and AI-generated content. This composition facilitates both quantitative and qualitative evaluation of interactive control accuracy and generalization. To demonstrate cross-scenario adaptability, we present exemplar results from diverse contexts.
% , like third-person gaming where the system autonomously adjusts cinematographic parameters in response to dynamic gameplay events.

\noindent \textbf{Evaluation Metrics.}
We employ several metrics for comprehensive evaluation to ensure fair comparison. We utilize Fréchet Video Distance(FVD)~\cite{unterthiner2019fvd} to evaluate the video realism. Relative pose error (RPE trans and RPE rot) are adopted to evaluate interactive control performance, after applying a Sim3 Umeyama alignment on the reconstructed trajectory of prediction to the ground truth. Following Matrix-Game, we employ Image Quality and Aesthetic scores for visual quality assessment, while utilizing Temporal Consistency to evaluate the visual and cinematographic continuity of generated sequences. For dynamic performance evaluation, we adapt the Dynamic Degree metric from VBench~\cite{huang2024vbench}, modifying its original binary classification approach to directly report absolute optical flow values as Dynamic Average, enabling a more nuanced, continuous assessment of motion characteristics. Additionally, we incorporate user preference scores obtained from user studies.
% z, as detailed in Sec.~\ref{sec:comp}.

\noindent\textbf{Baselines.} We compare our method with four representative baselines, including a current state-of-the-art open-sourced interactive game model, Matrix-Game, and three camera-controlled generation works: CameraCtrl~\cite{he2024cameractrl}, MotionCtrl~\cite{wang2024motionctrl} and WanX-Cam~\cite{wan2025}. The CameraCtrl and MotionCtrl employ the image-to-video SVD implementation, while WanX-Cam corresponds to the VideoX-Fun implementation.

\begin{table*}[!ht]\footnotesize
\centering
% \caption{Multi-dimensional Evaluation of Different Models}
\renewcommand{\arraystretch}{0.9}

\label{tab:model_eval}
\begin{tabular}{l|cccc cc c c} % S列自动对齐小数
\toprule
\multirow{2}{*}{Model} & 
\multicolumn{4}{c}{Visual Quality} & 
\multicolumn{1}{c}{Temporal} &
\multicolumn{2}{c}{RPE} &
\multirow{2}{*}{\thead{Infer Speed$\uparrow$ \\ (FPS)}}
\\
\cmidrule(lr){2-5} \cmidrule(lr){6-6} \cmidrule(lr){7-8}
& {FVD$\downarrow$} & {Image Quality$\uparrow$} & {Dynamic Average$\uparrow$} & {Aesthetic$\uparrow$} & {Temporal Consistency$\uparrow$}  & {Trans$\downarrow$} & {Rot$\downarrow$} \\
\midrule
CameraCtrl & 1580.9 & 0.66 & 7.2 & 0.64 & 0.92  & 0.13 & 0.25 & 1.75\\
MotionCtrl & 1902.0 & 0.68 & 7.8 & 0.48 & 0.94  & 0.17 & 0.32 & 0.67\\
WanX-Cam   & 1677.6 & 0.70 & 17.8 & \textbf{0.67} & 0.92  & 0.16 & 0.36 & 0.13 \\
\midrule
Matrix-Game & 2260.7 & \textbf{0.72} & 31.7 & 0.65 & 0.94  & 0.18 & 0.35 & 0.06 \\
\textbf{Ours} & \textbf{1554.2} & 0.69 & \textbf{67.2} & \textbf{0.67} & \textbf{0.95}  & \textbf{0.08} & \textbf{0.20} & 0.25\\
\textbf{Ours + PCM} & 1883.3 & 0.67 & 43.8 & 0.65 & 0.93  & \textbf{0.08} & \textbf{0.20}  &  \textbf{6.6}\\
\bottomrule
\end{tabular}
\caption{Quantitative comparison with recent related works. $\uparrow$ indicates higher is better, while $\downarrow$ indicates that lower is better. The best result is shown in \textbf{bold}.}
\label{tab:model_eval}
\end{table*}

\if0
\begin{table*}[!t]\footnotesize
\centering
% \vspace{-4pt}
\setlength{\tabcolsep}{1.5pt}
\label{tab:comparsion}
\begin{tabular}{l|cccccccc}
\toprule
% \multicolumn{3}{c}{Visual Quality} & 
% \multicolumn{2}{c}{Temporal Quality} &
% \multicolumn{2}{c}{Action Controllability} &
% \multirow{2}{*}{Memory}
% \\
% \cmidrule(lr){2-4} \cmidrule(lr){5-6} \cmidrule(lr){7-8}

& FVD & Image Quality$\uparrow$
 & Dynamic Average$\uparrow$ & Aesthetic$\uparrow$ & Temporal Cons.$\uparrow$ & Motion Smooth$\uparrow$ & RPE. trans$\downarrow$ & RPE. rot$\downarrow$    \\
\midrule                
CameraCtrl     & & 0.66          &  7.2        &     0.64    &    0.92       &   \textbf{0.98}  & 0.13 & 0.25              \\
MotionCtrl &  &   0.68        & 7.8        &        0.48   &   0.94      &       \textbf{ 0.98  }    & 0.17 & 0.32    \\
WanX-Cam &  & \textbf{0.70 }& 17.8 & \textbf{0.67} & 0.92 & \textbf{0.98} & 0.16 & 0.36\\
Ours     &       & 0.69 & \textbf{67.2} & \textbf{0.67} & \textbf{0.95} & \textbf{0.98} & \textbf{0.08} & \textbf{0.20} \\
\bottomrule
\end{tabular}
\caption{Quantitative comparisons with baselines. $\uparrow$ indicates higher is better, while $\downarrow$ indicates that lower is better. The best result is shown in \textbf{bold}.
 }
% \vspace{-18pt}
\end{table*}
\fi

\subsection{Comparisons with other methods}
\label{sec:comp}

\noindent \textbf{Quantitative Comparison.}
We conduct comprehensive comparisons with Matrix-Game, the current leading open-source game interaction model, under identical gaming scenarios. Despite employing the same base model~\cite{kong2024hunyuanvideo}, Hunyuan-GameCraft demonstrates significant improvements across the majority of key metrics, including generation quality, dynamic capability, control accuracy, and temporal consistency as shown in Tab.~\ref{tab:model_eval}. 
Notably, Hunyuan-GameCraft achieves the best results in dynamic performance compared to Matrix-Game, while simultaneously reducing interaction errors by $55\%$ in cross-domain tests. These advancements are attributable to our optimized training strategy and conditional injection mechanism, which collectively enable robust interactive generation across both gaming scenarios and diverse artistic styles.
% Most notably, we achieve XX\% performance gain in historical information retention, providing strong evidence for the effectiveness of our hybrid conditioning training approach. This advancement enables robust instruction following and content consistency even when processing highly varied action sequences with substantial magnitude changes. 

% GameCraft demonstrates several novel capabilities, including fine-grained velocity and angular control, along with powerful dynamic generation abilities. These advancements are enabled by our proposed continuous action space representation and optimized training data strategy. While certain aspects of these features are quantitatively captured by the XX and XX metrics, more intuitive demonstrations can be found in Section X.X and the supplementary video materials.

% Due to the bi-directional conversion property between continuous action space and camera trajectory, GameCraft can naturally extend to conventional camera-controlled video generation tasks. 
We also evaluate generation quality and control accuracy on the same test set, with quantitative results presented in Tab.~\ref{tab:model_eval}.
% While our model is primarily trained on gaming and rendered scenes, with only limited RealEstate10k data to enhance the controllability, 
Hunyuan-GameCraft demonstrates superior performance compared to other baselines.
The results suggest that our action-space formulation captures fundamental principles of camera motion that transcend game scene characteristics.
% This cross-domain effectiveness stems from our hybrid conditioning training paradigm and the continuous action injection mechanism, which jointly enable robust generalization across different scene types and motion patterns.
Furthermore, we report the inference speed of each baseline. Our method can achieve nearly real-time inference while slightly damaging the dynamic and visual quality, which is more suitable for game scene interaction.

\noindent \textbf{Qualitative Comparison.}
As shown in Fig.~\ref{fig:qualitative_compare}, we qualitatively demonstrate superior capabilities of Hunyuan-GameCraft from multiple perspectives. 
The part(a) compares our method with Matrix-Game in sequential single-action scenarios, using the Minecraft environment originally employed for training of Matrix-Game. The results demonstrate significantly superior interaction capabilities of Hunyuan-GameCraft. Furthermore, continuous left-right rotations effectively showcase the enhanced historical information retention enabled by hybrid history condition training approach.
The comparison of both game interaction models with sequential coupled action is shown in (b). Our method can accurately map input-coupled interaction signals while maintaining both quality consistency and spatial coherence during long video extension, achieving an immersive exploration experience. Part(c) focuses on evaluating image-to-video generation performance under single action across all baselines. Hunyuan-GameCraft demonstrates significant advantages in dynamic capability, including windmill rotation consistency, as well as overall visual quality.
% The first X rows compare it with Matrix-Game in gaming scenarios, evaluating both sequential single-action and coupled multi-action interactions. Across constrained environments like Minecraft and various AAA game settings, GameCraft demonstrates significantly more comprehensive interactive capabilities through our proposed continuous action space injection, including fundamental instruction following and effective fine-grained velocity/angle control. The X-th row showcases the long-term consistency and memory retention. Through hybrid historical conditioning training, GameCraft adaptively processes variable-length historical windows to achieve optimal scene information retention.

% Our approach further exhibits physically plausible dynamic generation capabilities, extending to specialized game genres as demonstrated in the X-th row's racing games, pedestrian NPCs, and third-person RPG scenarios. With carefully designed training data pipelines, balanced distribution strategies, and effective control injection, GameCraft represents the first open-source model compatible with both first-person camera control and third-person character manipulation. The final row demonstrates real-world scene adaptation and artistic style generalization. Benefiting from MMDiT's strong foundational capabilities, GameCraft achieves zero-shot transfer to unseen artistic styles (e.g., hand-drawn aesthetics) while maintaining functional interactivity, highlighting its general-purpose potential.

\noindent \textbf{User Study.}
Given the current lack of comprehensive benchmarks for interactive video generation models in both gaming and general scenarios, we conducted a user study involving 30 evaluators to enhance the reliability of our assessment. As shown in Tab.~\ref{tab:user}, our method achieved the highest scores by a margin across multiple dimensions in the anonymous user rankings.

\begin{table}[t]
\centering\footnotesize
\renewcommand{\arraystretch}{0.9}
\setlength{\tabcolsep}{0.7mm}{
\begin{tabular}{l|ccccc}
\toprule
 Method & \thead{Video \\ Quality}$\uparrow$ & \thead{Temporal\\ Consistency}$\uparrow$ & \thead{Motion \\ Smooth}$\uparrow$ & \thead{Action \\ Accuracy}$\uparrow$ & Dynamic$\uparrow$ \\ \midrule
CameraCtrl & 2.20 & 2.40 & 2.16 & 2.87 & 2.57 \\
MotionCtrl  & 3.23  & 3.20 & 3.21 & 3.09 & 3.22 \\
WanX-Cam &  2.42 & 2.53 & 2.44 & 2.81 & 2.46\\
Matrix-Game &  2.72 & 2.43 & 2.75 & 1.63 & 2.21\\
\midrule
Ours& \textbf{4.42} & \textbf{4.44} & \textbf{4.53} & \textbf{4.61} & \textbf{4.54}\\
\bottomrule
\end{tabular}}
\caption{Average ranking score of user study. For each object, users are asked to give a rank score where 5 for the best, and 1 for the worst. User prefer ours the best in both aspects. }
\label{tab:user}
\end{table}

\subsection{Ablation Study}
\label{sec:ab}
In this section, comprehensive experiments are conducted to validate the effectiveness of our contributions, including the data distribution, control injection, and hybrid history conditioning. 
% Each ablation study uses the same sequence of actions and reference image from the test set as input.

\begin{table}[!t]
\centering\footnotesize
\renewcommand{\arraystretch}{0.9}
\setlength{\tabcolsep}{1.2pt}
\begin{tabular}{lccccc}
\toprule
& FVD$\downarrow$ & DA$\uparrow$ & Aesthetic$\uparrow$ & RPE trans$\downarrow$ & RPE rot$\downarrow$             \\
\midrule                
(a) Only Synthetic Data          & 2550.7         & 34.6          & 0.56         & \textbf{0.07}       & \textbf{0.17} \\
(b) Only Live Data               & 1937.7        & \textbf{77.2 }      &   0.60        & 0.16          & 0.27 \\
\midrule                
(c) Token Concat.           & 2236.4        & 59.7          & 0.54        &  0.13        & 0.29 \\
(d) Channel-wise Concat.    &  1725.5         & 63.2        &     0.49       &  0.11        & 0.25 \\
\midrule                
(e) Image Condition       &  1655.3      & 47.6          & 0.58           & \textbf{0.07}       & 0.22 \\
(f) Clip Condition          & 1743.5         & 55.3         & 0.57          & 0.16       & 0.30 \\
\midrule   
(g) Ours (Render:Live=1:5)                   & \textbf{1554.2 }         & 67.2          & \textbf{0.67}    & 0.08 & 0.20 \\
\bottomrule
\end{tabular}
\caption{Ablation study on different data distribution, control injection, and hybrid history conditioning. DA denotes Dynamic Average score.}
\label{tab:ablation}
\end{table}

\noindent \textbf{Data Distribution.}
% Our full-parameter training of the base model prioritizes performance in low-noise gaming environments, emphasizing high-fidelity generation capabilities, which makes the training data distribution particularly crucial. 
% We analyze the data ratios 
To understand the distinct contributions of game data and synthetic data, we began with an ablation study evaluating their impact on the model's capabilities.
Notably, the synthetic data does not highlight dynamic objects due to the computational expense and complexity of generating dynamical scenes. Tab.~\ref{tab:ablation}(a)(b) demonstrate that training exclusively on synthetic data significantly improves interaction accuracy but substantially degrades dynamic generation capabilities, while gameplay data exhibits the opposite characteristics. Our training distribution achieves balanced results.
% between interaction accuracy and dynamic with distribution balancing and data augmentation technologies.

\noindent \textbf{Action Control Injection.}
Here we present ablation details for our camera injection experiments. Since the Plücker embeddings are already temporally and spatially aligned with the video latent representations, we implement three straightforward camera control schemes: (i) Token Addition, (ii) Token Concatenation, and (iii) Channel-wise Concatenation, as shown in the Tab.~\ref{tab:ablation}(c)(d)(g). Simply adding control signals at the initial stage achieves state-of-the-art control performance. Considering computational efficiency, we ultimately adopt Token Addition in our framework.

\noindent \textbf{Hybrid History Conditioning.}
Hunyuan-GameCraft implements hybrid history conditioning for video generation and extension. Fig.~\ref{fig:ab_long} visually demonstrates visual results under different conditioning schemes, while we provide quantitative ablation analysis here. As shown in Tab.~\ref{tab:ablation}(e)(f)(g), Hunyuan-GameCraft achieves satisfactory control accuracy when trained with single frame conditioning, yet suffers from quality degradation over multiple action sequences due to limited historical context, leading to quality collapse as shown in Fig.~\ref{fig:ab_long}. When employing historical clip conditioning, the model exhibits degraded interaction accuracy when processing control signals that significantly deviate from historical motions. Our hybrid history conditioning effectively balances this trade-off, enabling Hunyuan-GameCraft to simultaneously achieve superior interaction performance, long-term consistency and visual quality.

\begin{figure*}[t]
    \centering
    \includegraphics[width=\linewidth]{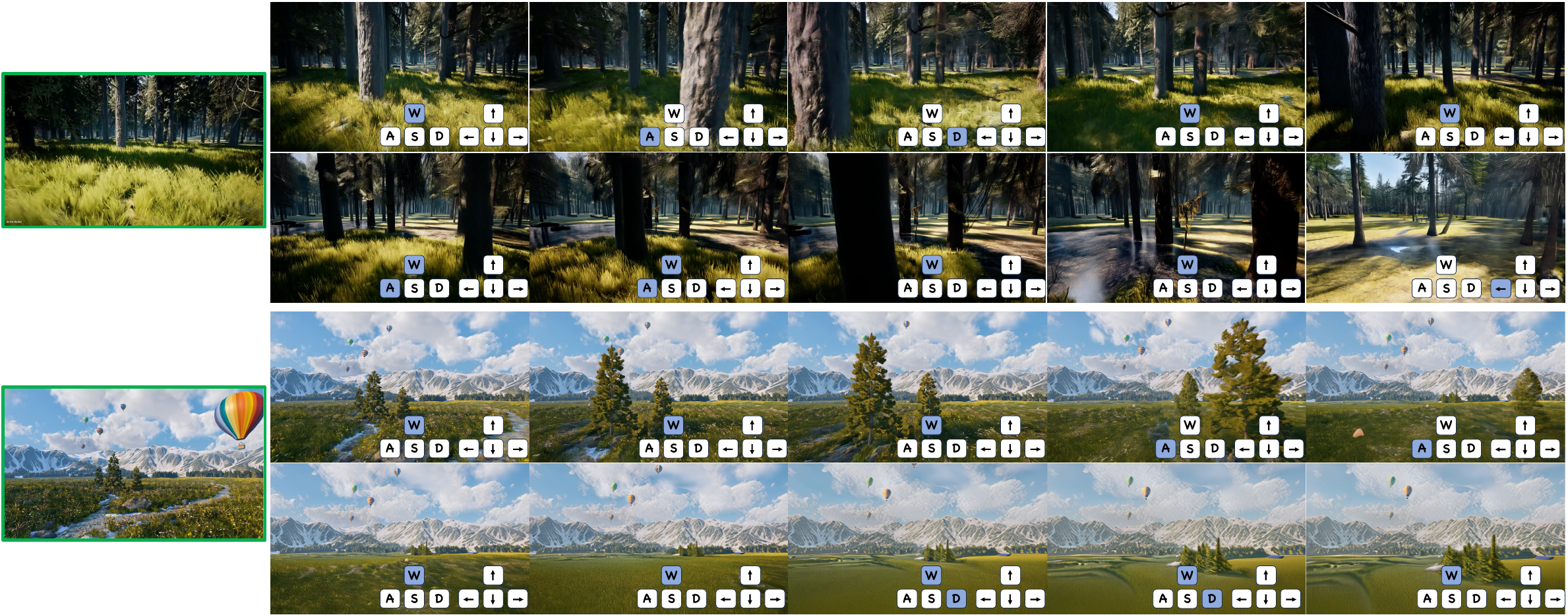}
    \caption{Long Video Extension Results. Hunyuan-GameCraft can generate minute-level video clips in length while maintaining the visual quality. }
    \label{fig:long_res}
\end{figure*}

\begin{figure}[t]
    \includegraphics[width=\linewidth]{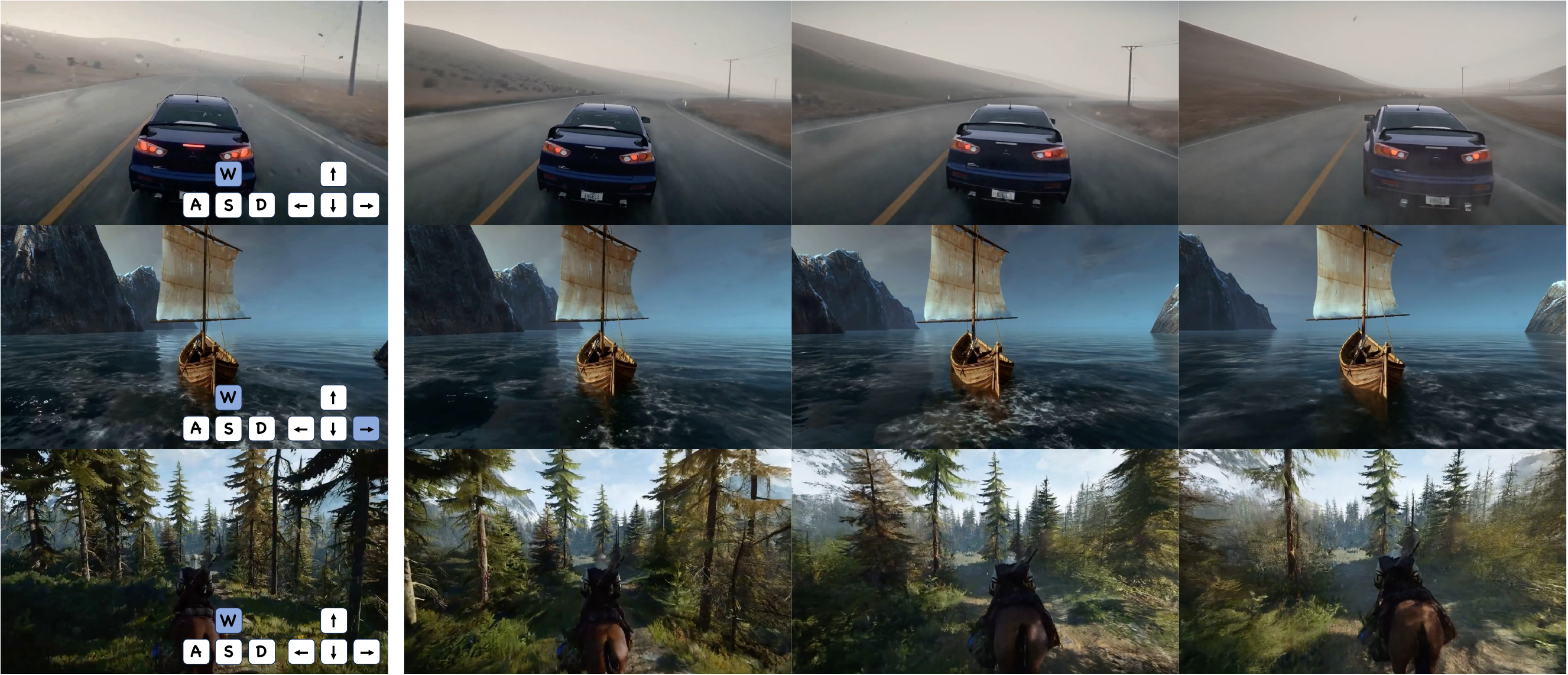}
    \caption{Interactive results on the third-perspective game video generation.}
    \label{fig:car}
\end{figure}

\begin{figure}[t]
    \includegraphics[width=\linewidth]{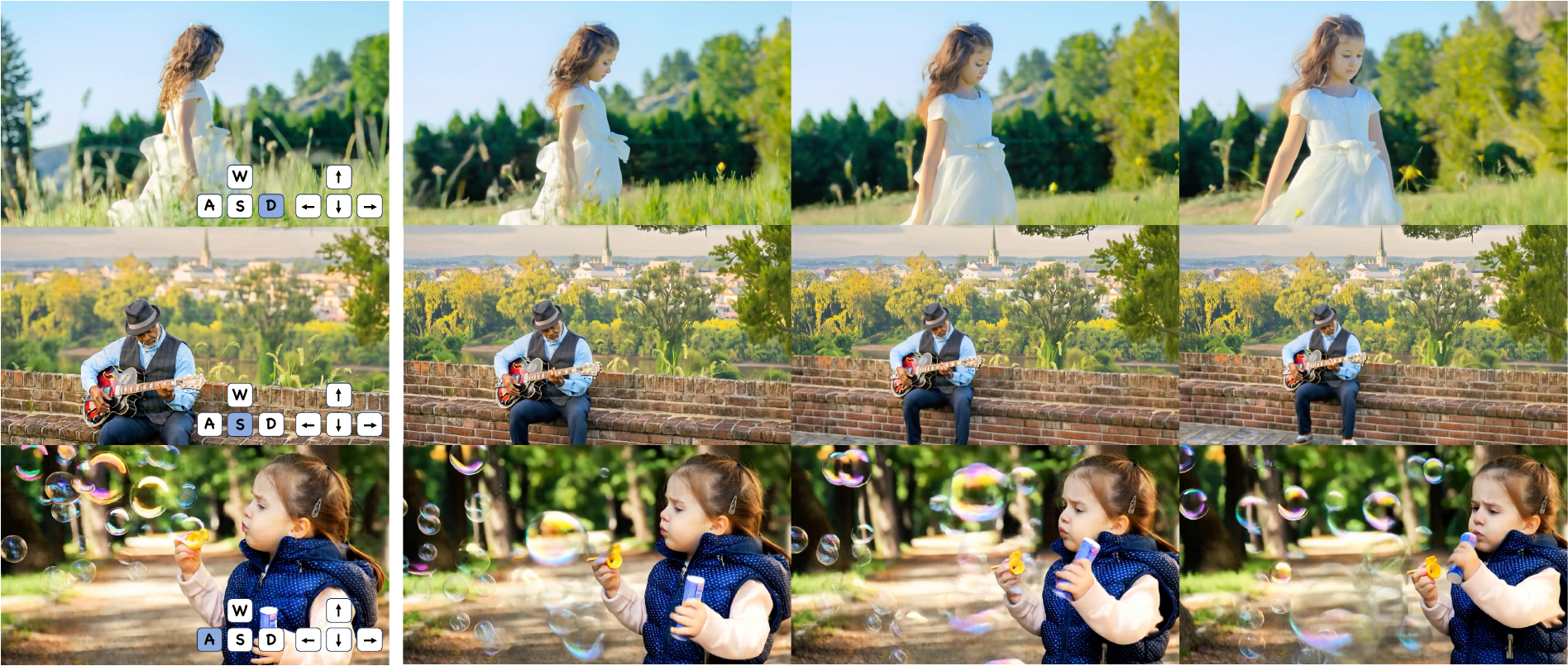}
    \caption{Hunyuan-GameCraft enables high-fidelity and high-dynamic real-world video generation with accurate camera control.}
    \label{fig:real}
\end{figure}

% \subsection{Novel Applications}

\section{Generalization on Real Worlds}
% \subsection{Generalization on Real Worlds}

Although our model is tailored for game scenes, the integration of a pre-trained video foundation model significantly enhances its generalization capabilities, enabling it to generate interactive videos in real-world domains as well. As shown in Fig~\ref{fig:real}, given images in real world, Hunyuan-GameCraft can successfully generate reasonable video with conditioned camera movement while keeping the dynamics. 

% \subsection{Video extension}

% \subsection{Memory}

\section{Limitations and Future Work}

While Hunyuan-GameCraft demonstrates impressive capabilities in interactive game video generation, its current action space is mainly tailored to open-world exploration and lacks a wider array of game-specific actions such as shooting, throwing, and explosions. In future work, we will expand the dataset with more diverse gameplay elements. Building on our advancements in controllability, long-form video generation, and history preservation, we will focus on developing the next-generation model for more physical and playable game interactions.

\section{Conclusion}

In this paper, we introduce Hunyuan-GameCraft, a significant step forward in interactive video generation. Through a unified action representation, hybrid history-conditioned training, and model distillation, our framework enables fine-grained control, efficient inference, and scalable long video synthesis. Besides, Hunyuan-GameCraft delivers enhanced realism, responsiveness, and temporal coherence. Our results demonstrate substantial improvements over existing methods, establishing Hunyuan-GameCraft as a robust foundation for future research and real-time deployment in immersive gaming environments.

% \newpage
{
    \small
    \bibliographystyle{ieeenat_fullname}
    \bibliography{main}

\begin{thebibliography}{36}
\providecommand{\natexlab}[1]{#1}
\providecommand{\url}[1]{\texttt{#1}}
\expandafter\ifx\csname urlstyle\endcsname\relax
  \providecommand{\doi}[1]{doi: #1}\else
  \providecommand{\doi}{doi: \begingroup \urlstyle{rm}\Url}\fi

\bibitem[Blattmann et~al.(2023{\natexlab{a}})Blattmann, Dockhorn, Kulal, Mendelevitch, Kilian, Lorenz, Levi, English, Voleti, Letts, et~al.]{blattmann2023stable}
Andreas Blattmann, Tim Dockhorn, Sumith Kulal, Daniel Mendelevitch, Maciej Kilian, Dominik Lorenz, Yam Levi, Zion English, Vikram Voleti, Adam Letts, et~al.
\newblock Stable video diffusion: Scaling latent video diffusion models to large datasets.
\newblock \emph{arXiv preprint arXiv:2311.15127}, 2023{\natexlab{a}}.

\bibitem[Blattmann et~al.(2023{\natexlab{b}})Blattmann, Rombach, Ling, Dockhorn, Kim, Fidler, and Kreis]{blattmann2023align}
Andreas Blattmann, Robin Rombach, Huan Ling, Tim Dockhorn, Seung~Wook Kim, Sanja Fidler, and Karsten Kreis.
\newblock Align your latents: High-resolution video synthesis with latent diffusion models.
\newblock In \emph{Proceedings of the IEEE/CVF conference on computer vision and pattern recognition}, pages 22563--22575, 2023{\natexlab{b}}.

\bibitem[Bradski(2000)]{bradski2000opencv}
Gary Bradski.
\newblock The opencv library.
\newblock \emph{Dr. Dobb's Journal: Software Tools for the Professional Programmer}, 25\penalty0 (11):\penalty0 120--123, 2000.

\bibitem[Castellano()]{Castellano_PySceneDetect}
Brandon Castellano.
\newblock {PySceneDetect}.

\bibitem[Che et~al.(2025)Che, He, Liu, Jin, and Chen]{che2024gamegen}
Haoxuan Che, Xuanhua He, Quande Liu, Cheng Jin, and Hao Chen.
\newblock Gamegen-x: Interactive open-world game video generation.
\newblock In \emph{International Conference on Learning Representations}, 2025.

\bibitem[Chen et~al.(2024)Chen, Mart{\'\i}~Mons{\'o}, Du, Simchowitz, Tedrake, and Sitzmann]{chen2024diffusion}
Boyuan Chen, Diego Mart{\'\i}~Mons{\'o}, Yilun Du, Max Simchowitz, Russ Tedrake, and Vincent Sitzmann.
\newblock Diffusion forcing: Next-token prediction meets full-sequence diffusion.
\newblock \emph{Advances in Neural Information Processing Systems}, 37:\penalty0 24081--24125, 2024.

\bibitem[Dalal et~al.(2025)Dalal, Koceja, Hussein, Xu, Zhao, Song, Han, Cheung, Kautz, Guestrin, et~al.]{dalal2025one}
Karan Dalal, Daniel Koceja, Gashon Hussein, Jiarui Xu, Yue Zhao, Youjin Song, Shihao Han, Ka~Chun Cheung, Jan Kautz, Carlos Guestrin, et~al.
\newblock One-minute video generation with test-time training.
\newblock \emph{arXiv preprint arXiv:2504.05298}, 2025.

\bibitem[Decard(2024)]{oasis}
Decard.
\newblock Oasis: A universe in a transformer.
\newblock \url{https://www.decart.ai/articles/oasis-interactive-ai-video-game-model}, 2024.

\bibitem[Esser et~al.(2024)Esser, Kulal, Blattmann, Entezari, M{\"u}ller, Saini, Levi, Lorenz, Sauer, Boesel, et~al.]{esser2024scaling}
Patrick Esser, Sumith Kulal, Andreas Blattmann, Rahim Entezari, Jonas M{\"u}ller, Harry Saini, Yam Levi, Dominik Lorenz, Axel Sauer, Frederic Boesel, et~al.
\newblock Scaling rectified flow transformers for high-resolution image synthesis.
\newblock In \emph{Forty-first international conference on machine learning}, 2024.

\bibitem[Feng et~al.(2024)Feng, Zhang, Yang, Xiao, Shu, Liu, Zheng, Huang, Liu, and Zhang]{feng2024matrix}
Ruili Feng, Han Zhang, Zhantao Yang, Jie Xiao, Zhilei Shu, Zhiheng Liu, Andy Zheng, Yukun Huang, Yu Liu, and Hongyang Zhang.
\newblock The matrix: Infinite-horizon world generation with real-time moving control.
\newblock \emph{arXiv preprint arXiv:2412.03568}, 2024.

\bibitem[Gao et~al.(2024)Gao, Shi, Zhang, Wang, and Xiao]{gao2024vid}
Kaifeng Gao, Jiaxin Shi, Hanwang Zhang, Chunping Wang, and Jun Xiao.
\newblock Vid-gpt: Introducing gpt-style autoregressive generation in video diffusion models.
\newblock \emph{arXiv preprint arXiv:2406.10981}, 2024.

\bibitem[Gu et~al.(2025)Gu, Mao, and Shou]{gu2025long}
Yuchao Gu, Weijia Mao, and Mike~Zheng Shou.
\newblock Long-context autoregressive video modeling with next-frame prediction.
\newblock \emph{arXiv preprint arXiv:2503.19325}, 2025.

\bibitem[He et~al.(2024)He, Xu, Guo, Wetzstein, Dai, Li, and Yang]{he2024cameractrl}
Hao He, Yinghao Xu, Yuwei Guo, Gordon Wetzstein, Bo Dai, Hongsheng Li, and Ceyuan Yang.
\newblock Cameractrl: Enabling camera control for text-to-video generation.
\newblock \emph{arXiv preprint arXiv:2404.02101}, 2024.

\bibitem[He et~al.(2025)He, Yang, Lin, Xu, Wei, Gui, Zhao, Wetzstein, Jiang, and Li]{he2025cameractrl}
Hao He, Ceyuan Yang, Shanchuan Lin, Yinghao Xu, Meng Wei, Liangke Gui, Qi Zhao, Gordon Wetzstein, Lu Jiang, and Hongsheng Li.
\newblock Cameractrl ii: Dynamic scene exploration via camera-controlled video diffusion models.
\newblock \emph{arXiv preprint arXiv:2503.10592}, 2025.

\bibitem[Henschel et~al.(2024)Henschel, Khachatryan, Poghosyan, Hayrapetyan, Tadevosyan, Wang, Navasardyan, and Shi]{henschel2024streamingt2v}
Roberto Henschel, Levon Khachatryan, Hayk Poghosyan, Daniil Hayrapetyan, Vahram Tadevosyan, Zhangyang Wang, Shant Navasardyan, and Humphrey Shi.
\newblock Streamingt2v: Consistent, dynamic, and extendable long video generation from text.
\newblock \emph{arXiv preprint arXiv:2403.14773}, 2024.

\bibitem[Huang et~al.(2024)Huang, He, Yu, Zhang, Si, Jiang, Zhang, Wu, Jin, Chanpaisit, et~al.]{huang2024vbench}
Ziqi Huang, Yinan He, Jiashuo Yu, Fan Zhang, Chenyang Si, Yuming Jiang, Yuanhan Zhang, Tianxing Wu, Qingyang Jin, Nattapol Chanpaisit, et~al.
\newblock Vbench: Comprehensive benchmark suite for video generative models.
\newblock In \emph{Proceedings of the IEEE/CVF Conference on Computer Vision and Pattern Recognition}, pages 21807--21818, 2024.

\bibitem[KolorsTeam(2024)]{kolors}
KolorsTeam.
\newblock Kolors: Effective training of diffusion model for photorealistic text-to-image synthesis.
\newblock \emph{arXiv preprint}, 2024.

\bibitem[Kong et~al.(2024)Kong, Tian, Zhang, Min, Dai, Zhou, Xiong, Li, Wu, Zhang, et~al.]{kong2024hunyuanvideo}
Weijie Kong, Qi Tian, Zijian Zhang, Rox Min, Zuozhuo Dai, Jin Zhou, Jiangfeng Xiong, Xin Li, Bo Wu, Jianwei Zhang, et~al.
\newblock Hunyuanvideo: A systematic framework for large video generative models.
\newblock \emph{arXiv preprint arXiv:2412.03603}, 2024.

\bibitem[Li et~al.(2025)Li, Zhou, Zheng, Lu, Huang, Chen, Tang, Xu, Tao, Wang, et~al.]{li2025hunyuan}
Ruihuang Li, Caijin Zhou, Shoujian Zheng, Jianxiang Lu, Jiabin Huang, Comi Chen, Junshu Tang, Guangzheng Xu, Jiale Tao, Hongmei Wang, et~al.
\newblock Hunyuan-game: Industrial-grade intelligent game creation model.
\newblock \emph{arXiv preprint arXiv:2505.14135}, 2025.

\bibitem[Lu et~al.(2024)Lu, Liang, Zhu, and Yang]{lu2024freelong}
Yu Lu, Yuanzhi Liang, Linchao Zhu, and Yi Yang.
\newblock Freelong: Training-free long video generation with spectralblend temporal attention.
\newblock \emph{arXiv preprint arXiv:2407.19918}, 2024.

\bibitem[Luo et~al.(2023)Luo, Tan, Huang, Li, and Zhao]{luo2023latent}
Simian Luo, Yiqin Tan, Longbo Huang, Jian Li, and Hang Zhao.
\newblock Latent consistency models: Synthesizing high-resolution images with few-step inference.
\newblock \emph{arXiv preprint arXiv:2310.04378}, 2023.

\bibitem[Parker-Holder et~al.(2024)Parker-Holder, Ball, Bruce, Dasagi, Holsheimer, Kaplanis, Moufarek, Scully, Shar, Shi, Spencer, Yung, Dennis, Kenjeyev, Long, Mnih, Chan, Gazeau, Li, Pardo, Wang, Zhang, Besse, Harley, Mitenkova, Wang, Clune, Hassabis, Hadsell, Bolton, Singh, and Rockt{\"a}schel]{parkerholder2024genie2}
Jack Parker-Holder, Philip Ball, Jake Bruce, Vibhavari Dasagi, Kristian Holsheimer, Christos Kaplanis, Alexandre Moufarek, Guy Scully, Jeremy Shar, Jimmy Shi, Stephen Spencer, Jessica Yung, Michael Dennis, Sultan Kenjeyev, Shangbang Long, Vlad Mnih, Harris Chan, Maxime Gazeau, Bonnie Li, Fabio Pardo, Luyu Wang, Lei Zhang, Frederic Besse, Tim Harley, Anna Mitenkova, Jane Wang, Jeff Clune, Demis Hassabis, Raia Hadsell, Adrian Bolton, Satinder Singh, and Tim Rockt{\"a}schel.
\newblock Genie 2: A large-scale foundation world model.
\newblock 2024.

\bibitem[Skorokhodov et~al.(2022)Skorokhodov, Tulyakov, and Elhoseiny]{skorokhodov2022stylegan}
Ivan Skorokhodov, Sergey Tulyakov, and Mohamed Elhoseiny.
\newblock Stylegan-v: A continuous video generator with the price, image quality and perks of stylegan2.
\newblock In \emph{Proceedings of the IEEE/CVF conference on computer vision and pattern recognition}, pages 3626--3636, 2022.

\bibitem[Teed and Deng(2020)]{teed2020raft}
Zachary Teed and Jia Deng.
\newblock Raft: Recurrent all-pairs field transforms for optical flow.
\newblock In \emph{Computer Vision--ECCV 2020: 16th European Conference, Glasgow, UK, August 23--28, 2020, Proceedings, Part II 16}, pages 402--419. Springer, 2020.

\bibitem[Unterthiner et~al.(2019)Unterthiner, Van~Steenkiste, Kurach, Marinier, Michalski, and Gelly]{unterthiner2019fvd}
Thomas Unterthiner, Sjoerd Van~Steenkiste, Karol Kurach, Rapha{\"e}l Marinier, Marcin Michalski, and Sylvain Gelly.
\newblock Fvd: A new metric for video generation.
\newblock 2019.

\bibitem[Valevski et~al.(2024)Valevski, Leviathan, Arar, and Fruchter]{valevski2024diffusion}
Dani Valevski, Yaniv Leviathan, Moab Arar, and Shlomi Fruchter.
\newblock Diffusion models are real-time game engines.
\newblock \emph{arXiv preprint arXiv:2408.14837}, 2024.

\bibitem[Wan et~al.(2025)Wan, Wang, Ai, Wen, Mao, Xie, Chen, Yu, Zhao, Yang, Zeng, Wang, Zhang, Zhou, Wang, Chen, Zhu, Zhao, Yan, Huang, Feng, Zhang, Li, Wu, Chu, Feng, Zhang, Sun, Fang, Wang, Gui, Weng, Shen, Lin, Wang, Wang, Zhou, Wang, Shen, Yu, Shi, Huang, Xu, Kou, Lv, Li, Liu, Wang, Zhang, Huang, Li, Wu, Liu, Pan, Zheng, Hong, Shi, Feng, Jiang, Han, Wu, and Liu]{wan2025}
Team Wan, Ang Wang, Baole Ai, Bin Wen, Chaojie Mao, Chen-Wei Xie, Di Chen, Feiwu Yu, Haiming Zhao, Jianxiao Yang, Jianyuan Zeng, Jiayu Wang, Jingfeng Zhang, Jingren Zhou, Jinkai Wang, Jixuan Chen, Kai Zhu, Kang Zhao, Keyu Yan, Lianghua Huang, Mengyang Feng, Ningyi Zhang, Pandeng Li, Pingyu Wu, Ruihang Chu, Ruili Feng, Shiwei Zhang, Siyang Sun, Tao Fang, Tianxing Wang, Tianyi Gui, Tingyu Weng, Tong Shen, Wei Lin, Wei Wang, Wei Wang, Wenmeng Zhou, Wente Wang, Wenting Shen, Wenyuan Yu, Xianzhong Shi, Xiaoming Huang, Xin Xu, Yan Kou, Yangyu Lv, Yifei Li, Yijing Liu, Yiming Wang, Yingya Zhang, Yitong Huang, Yong Li, You Wu, Yu Liu, Yulin Pan, Yun Zheng, Yuntao Hong, Yupeng Shi, Yutong Feng, Zeyinzi Jiang, Zhen Han, Zhi-Fan Wu, and Ziyu Liu.
\newblock Wan: Open and advanced large-scale video generative models.
\newblock \emph{arXiv preprint arXiv:2503.20314}, 2025.

\bibitem[Wang et~al.(2024{\natexlab{a}})Wang, Huang, Bergman, Shen, Gao, Lingelbach, Sun, Bian, Song, Liu, et~al.]{wang2024phased}
Fu-Yun Wang, Zhaoyang Huang, Alexander Bergman, Dazhong Shen, Peng Gao, Michael Lingelbach, Keqiang Sun, Weikang Bian, Guanglu Song, Yu Liu, et~al.
\newblock Phased consistency models.
\newblock \emph{Advances in neural information processing systems}, 37:\penalty0 83951--84009, 2024{\natexlab{a}}.

\bibitem[Wang et~al.(2024{\natexlab{b}})Wang, Bai, Tan, Wang, Fan, Bai, Chen, Liu, Wang, Ge, et~al.]{wang2024qwen2}
Peng Wang, Shuai Bai, Sinan Tan, Shijie Wang, Zhihao Fan, Jinze Bai, Keqin Chen, Xuejing Liu, Jialin Wang, Wenbin Ge, et~al.
\newblock Qwen2-vl: Enhancing vision-language model's perception of the world at any resolution.
\newblock \emph{arXiv preprint arXiv:2409.12191}, 2024{\natexlab{b}}.

\bibitem[Wang et~al.(2024{\natexlab{c}})Wang, Zhu, Huang, Wang, Chen, and Lu]{wang2024worlddreamer}
Xiaofeng Wang, Zheng Zhu, Guan Huang, Boyuan Wang, Xinze Chen, and Jiwen Lu.
\newblock Worlddreamer: Towards general world models for video generation via predicting masked tokens.
\newblock \emph{arXiv preprint arXiv:2401.09985}, 2024{\natexlab{c}}.

\bibitem[Wang et~al.(2024{\natexlab{d}})Wang, Yuan, Wang, Li, Chen, Xia, Luo, and Shan]{wang2024motionctrl}
Zhouxia Wang, Ziyang Yuan, Xintao Wang, Yaowei Li, Tianshui Chen, Menghan Xia, Ping Luo, and Ying Shan.
\newblock Motionctrl: A unified and flexible motion controller for video generation.
\newblock In \emph{ACM SIGGRAPH 2024 Conference Papers}, pages 1--11, 2024{\natexlab{d}}.

\bibitem[WorldLabs(2024)]{worldlabs2024}
WorldLabs.
\newblock Generating worlds.
\newblock \url{https://www.worldlabs.ai/blog}, 2024.

\bibitem[Yang et~al.(2021)Yang, Liu, Chen, Shen, Hao, and Wang]{yang2021causalvae}
Mengyue Yang, Furui Liu, Zhitang Chen, Xinwei Shen, Jianye Hao, and Jun Wang.
\newblock Causalvae: Disentangled representation learning via neural structural causal models.
\newblock In \emph{Proceedings of the IEEE/CVF conference on computer vision and pattern recognition}, pages 9593--9602, 2021.

\bibitem[Yu et~al.(2025)Yu, Qin, Wang, Wan, Zhang, and Liu]{yu2025gamefactory}
Jiwen Yu, Yiran Qin, Xintao Wang, Pengfei Wan, Di Zhang, and Xihui Liu.
\newblock Gamefactory: Creating new games with generative interactive videos.
\newblock \emph{arXiv preprint arXiv:2501.08325}, 2025.

\bibitem[Zhang et~al.(2024)Zhang, Herrmann, Hur, Jampani, Darrell, Cole, Sun, and Yang]{zhang2024monst3r}
Junyi Zhang, Charles Herrmann, Junhwa Hur, Varun Jampani, Trevor Darrell, Forrester Cole, Deqing Sun, and Ming-Hsuan Yang.
\newblock Monst3r: A simple approach for estimating geometry in the presence of motion.
\newblock \emph{arXiv preprint arXiv:2410.03825}, 2024.

\bibitem[Zhang et~al.(2025)Zhang, Peng, Wang, Wang, Zhu, Gao, Li, Liu, and Zhou]{zhang2025matrixgame}
Yifan Zhang, Chunli Peng, Boyang Wang, Puyi Wang, Qingcheng Zhu, Zedong Gao, Eric Li, Yang Liu, and Yahui Zhou.
\newblock Matrix-game: Interactive world foundation model.
\newblock \emph{arXiv}, 2025.

\end{thebibliography}
}


\end{document}